%% file: vot2016submission.tex
\documentclass[runningheads]{llncs}
\usepackage{graphicx}
\usepackage{amsmath,amssymb} 
\usepackage{xcolor}
\usepackage[width=122mm,left=12mm,paperwidth=146mm,height=193mm,top=12mm,paperheight=217mm]{geometry}
\usepackage{enumitem}
\usepackage{array}
\usepackage{bbm}  
\usepackage{xspace}
\usepackage[colorlinks]{hyperref}
\usepackage{caption}
\usepackage{subcaption}
\newcolumntype{L}[1]{>{\raggedright\let\newline\\\arraybackslash\hspace{0pt}}m{#1}}
\newcolumntype{C}[1]{>{\centering\let\newline\\\arraybackslash\hspace{0pt}}m{#1}}
\newcolumntype{R}[1]{>{\raggedleft\let\newline\\\arraybackslash\hspace{0pt}}m{#1}}
\begin{document}
\pagestyle{headings}
\mainmatter

\input{macros}
\title{Fully-Convolutional Siamese Networks \\ for Object Tracking}

\titlerunning{Fully-Convolutional Siamese Networks for Object Tracking}

\authorrunning{L.\ Bertinetto, J.\ Valmadre, J.\ F.\ Henriques, A.\ Vedaldi, P.\ H.\ S.\ Torr}

\author{
Luca Bertinetto
\thanks{The first two authors contributed equally, and are listed in alphabetical order.}
\quad Jack Valmadre $ ^{\star}$
\quad Jo\~{a}o F.\ Henriques \\
\quad Andrea Vedaldi
\quad Philip H.\ S.\ Torr \\
}

\institute{Department of Engineering Science, University of Oxford \\ \texttt{\{name.surname\}@eng.ox.ac.uk}}

\maketitle

\begin{abstract}
The problem of arbitrary object tracking has traditionally been tackled by learning a model of the object's appearance exclusively online, using as sole training data the video itself.
Despite the success of these methods, their online-only approach inherently limits the richness of the model they can learn.
Recently, several attempts have been made to exploit the expressive power of deep convolutional networks.
However, when the object to track is not known beforehand, it is necessary to perform Stochastic Gradient Descent online to adapt the weights of the network, severely compromising the speed of the system.
In this paper we equip a basic tracking algorithm with a novel fully-convolutional Siamese network trained end-to-end on the ILSVRC15 dataset for object detection in video.
Our tracker operates at frame-rates beyond real-time and, despite its extreme simplicity, achieves state-of-the-art performance in multiple benchmarks.
\end{abstract}

\section{Introduction}
\input{introduction}

\section{Deep similarity learning for tracking}
\label{sec:methodology}
\input{methodology}

\section{Related work}
\input{related}

\section{Experiments}

\input{experiments}

\section{Conclusion}
In this work, we depart from the traditional online learning methodology employed in tracking, and show an alternative approach that focuses on learning strong embeddings in an offline phase. Differently from their use in classification settings, we demonstrate that for tracking applications Siamese fully-convolutional deep networks have the ability to use the available data more efficiently. This is reflected both at test-time, by performing efficient spatial searches, but also at training-time, where every sub-window effectively represents a useful sample with little extra cost. The experiments show that deep embeddings provide a naturally rich source of features for online trackers, and enable simplistic test-time strategies to perform well.
We believe that this approach is complementary to more sophisticated online tracking methodologies, and expect future work to explore this relationship more thoroughly.

\paragraph{Acknowledgments.} We are grateful for support by ERC StG 638009-IDIU.

\clearpage

\bibliographystyle{splncs}
\bibliography{vot2016submission}
\end{document}

%% file: macros.tex
\newcommand{\jack}[1]{{\color{blue} [Jack: #1]}}
\newcommand{\luca}[1]{{\color{orange} [Luca: #1]}}
\newcommand{\joao}[1]{{\color{green!50!black} [Joao: #1]}}
\newcommand{\todo}[1]{{\color{red} [TODO: #1]}}
\newcommand{\tocite}[1]{{\color{purple} [cite: #1]}}

\newcommand{\eg}{e.g.\xspace}
\newcommand{\ie}{i.e.\xspace}

\newcommand{\R}{\mathbb{R}}

\renewcommand{\paragraph}[1]{\par\noindent\textbf{#1}}

%% file: introduction.tex
We consider the problem of tracking an arbitrary object in video, where the object is identified solely by a rectangle in the first frame.
Since the algorithm may be requested to track any arbitrary object, it is impossible to have already gathered data and trained a specific detector.


For several years, the most successful paradigm for this scenario has been to learn a model of the object's appearance in an online fashion using examples extracted from the video itself~\cite{smeulders2014visual}.
This owes in large part to the demonstrated ability of methods like TLD~\cite{kalal2012tracking}, Struck~\cite{hare2011struck} and KCF~\cite{henriques2015high}.
However, a clear deficiency of using data derived exclusively from the current video is that only comparatively simple models can be learnt.
While other problems in computer vision have seen an increasingly pervasive adoption of deep convolutional networks (conv-nets) trained from large supervised datasets, the scarcity of supervised data and the constraint of real-time operation prevent the naive application of deep learning within this paradigm of learning a detector per video.

Several recent works have aimed to overcome this limitation using a pre-trained deep conv-net that was learnt for a different but related task.
These approaches either apply ``shallow'' methods (\eg correlation filters) using the network's internal representation as features~\cite{ma2015hierarchical,danelljan2015convolutional} or perform SGD (stochastic gradient descent) to fine-tune multiple layers of the network~\cite{wang2015transferring,wang2015visual,nam2015learning}.
While the use of shallow methods does not take full advantage of the benefits of end-to-end learning, methods that apply SGD during tracking to achieve state-of-the-art results have not been able to operate in real-time.

We advocate an alternative approach in which a deep conv-net is trained to address a more general \emph{similarity learning} problem in an initial offline phase, and then this function is simply evaluated online during tracking.
The key contribution of this paper is to demonstrate that this approach achieves very competitive performance in modern tracking benchmarks at speeds that far exceed the frame-rate requirement.
Specifically, we train a Siamese network to locate an \emph{exemplar} image within a larger \emph{search} image.
A further contribution is a novel Siamese architecture that is \emph{fully-convolutional} with respect to the search image: dense and efficient sliding-window evaluation is achieved with a bilinear layer that computes the cross-correlation of its two inputs.

We posit that the similarity learning approach has gone relatively neglected because the tracking community did not have access to vast labelled datasets.
In fact, until recently the available datasets comprised only a few hundred annotated videos.
However, we believe that the emergence of the ILSVRC dataset for object detection in video~\cite{ILSVRC15} (henceforth ImageNet Video) makes it possible to train such a model.
Furthermore, the fairness of training and testing deep models for tracking using videos from the same domain is a point of controversy, as it has been recently prohibited by the VOT committee.
We show that our model generalizes from the ImageNet Video domain to the ALOV/OTB/VOT~\cite{smeulders2014visual,WuLimYang13,kristan2015visual} domain, enabling the videos of tracking benchmarks to be reserved for testing purposes.


%% file: methodology.tex
Learning to track arbitrary objects can be addressed using similarity learning.
We propose to learn a function $f(z, x)$ that compares an exemplar image~$z$ to a candidate image~$x$ of the same size and returns a high score if the two images depict the same object and a low score otherwise.
To find the position of the object in a new image, we can then exhaustively test all possible locations and choose the candidate with the maximum similarity to the past appearance of the object.
In experiments, we will simply use the initial appearance of the object as the exemplar.
The function $f$ will be learnt from a dataset of videos with labelled object trajectories.

Given their widespread success in computer vision~\cite{razavian2014cnn,parkhi2015deep,dosovitskiy2015flownet,krizhevsky2012imagenet}, we will use a deep conv-net as the function~$f$.
Similarity learning with deep conv-nets is typically addressed using Siamese architectures~\cite{bromley1993signature,taigman2014deepface,zagoruyko2015learning}.
Siamese networks apply an identical transformation $\varphi$ to both inputs and then combine their representations using another function $g$ according to $f(z, x) = g(\varphi(z), \varphi(x))$.
When the function $g$ is a simple distance or similarity metric, the function $\varphi$ can be considered an embedding.
Deep Siamese conv-nets have previously been applied to tasks such as face verification~\cite{taigman2014deepface,schroff2015facenet,parkhi2015deep}, keypoint descriptor learning~\cite{zagoruyko2015learning,simo2015discriminative} and one-shot character recognition~\cite{koch2015siamese}.

\subsection{Fully-convolutional Siamese architecture}

We propose a Siamese architecture which is \emph{fully-convolutional} with respect to the candidate image~$x$.
We say that a function is fully-convolutional if it commutes with translation.
To give a more precise definition, introducing $L_{\tau}$ to denote the translation operator $(L_{\tau} x)[u] = x[u-\tau]$, a function $h$ that maps signals to signals is fully-convolutional with integer stride $k$ if
\begin{equation}\label{eq:commutes-translation}
h(L_{k \tau} x) = L_{\tau} h(x)
\end{equation}
for any translation~$\tau$.
(When $x$ is a finite signal, this only need hold for the valid region of the output.)

\begin{figure}[t]
\centering
\includegraphics[scale=.9]{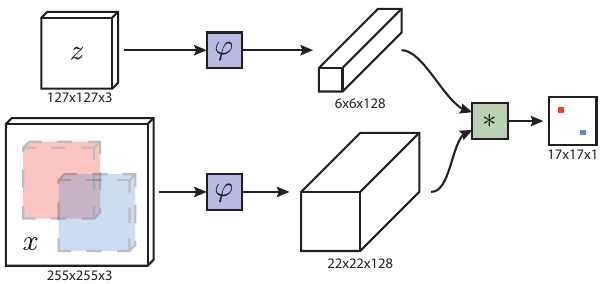}
\caption{Fully-convolutional Siamese architecture.
Our architecture is fully-convolutional with respect to the search image~$x$.
The output is a scalar-valued score map whose dimension depends on the size of the search image.
This enables the similarity function to be computed for all translated sub-windows within the search image in one evaluation.
In this example, the red and blue pixels in the score map contain the similarities for the corresponding sub-windows.
Best viewed in colour.
}
\label{fig:fully-conv-siamese}
\end{figure}

The advantage of a fully-convolutional network is that, instead of a candidate image of the same size, we can provide as input to the network a much larger \emph{search} image and it will compute the similarity at all translated sub-windows on a dense grid in a single evaluation.
To achieve this, we use a convolutional embedding function $\varphi$ and combine the resulting feature maps using a cross-correlation layer
\begin{equation}\label{eq:cross-correlation}
f(z, x) = \varphi(z) * \varphi(x) + b \, \mathbbm{1},
\end{equation}
where $b \, \mathbbm{1}$ denotes a signal which takes value $b \in \R$ in every location.
The output of this network is not a single score but rather a score map defined on a finite grid~$\mathcal{D} \subset \mathbb{Z}^{2}$ as illustrated in Figure~\ref{fig:fully-conv-siamese}.
Note that the output of the embedding function is a feature map with spatial support as opposed to a plain vector.
The same technique has been applied in contemporary work on stereo matching~\cite{luo2016efficient}.

During tracking, we use a search image centred at the previous position of the target.
The position of the maximum score relative to the centre of the score map, multiplied by the stride of the network, gives the displacement of the target from frame to frame.
Multiple scales are searched in a single forward-pass by assembling a mini-batch of scaled images.

Combining feature maps using cross-correlation and evaluating the network once on the larger search image is mathematically equivalent to combining feature maps using the inner product and evaluating the network on each translated sub-window independently.
However, the cross-correlation layer provides an incredibly simple method to implement this operation efficiently within the framework of existing conv-net libraries.
While this is clearly useful during testing, it can also be exploited during training.

\subsection{Training with large search images\label{sec:training}}

We employ a discriminative approach, training the network on positive and negative pairs and adopting the logistic loss
\begin{equation}
\ell(y, v) = \log(1 + \exp(-y v))
\end{equation}
where $v$ is the real-valued score of a single exemplar-candidate pair and $y \in \{+1, -1\}$ is its ground-truth label.
We exploit the fully-convolutional nature of our network during training by using pairs that comprise an exemplar image and a larger search image.
This will produce a map of scores $v : \mathcal{D} \to \R$, effectively generating many examples per pair.
We define the loss of a score map to be the mean of the individual losses
\begin{equation}\label{eq:global-loss}
L(y, v) = \frac{1}{|\mathcal{D}|} \sum_{u \in \mathcal{D}} \ell(y[u], v[u]) \enspace ,
\end{equation}
requiring a true label $y[u] \in \{+1, -1\}$ for each position $u \in \mathcal{D}$ in the score map.
The parameters of the conv-net $\theta$ are obtained by applying Stochastic Gradient Descent (SGD) to the problem
\begin{equation}
\arg\min_{\theta} \underset{(z, x, y)}{\mathbb{E}} L(y, f(z, x; \theta)) \enspace .
\label{eq:training}
\end{equation}

\begin{figure}[t]
\centering
\input{art/pairs/figure}
\caption{
Training pairs extracted from the same video: exemplar image and corresponding search image from same video.
When a sub-window extends beyond the extent of the image, the missing portions are filled with the mean RGB value.
}
\label{fig:pos-pairs}
\end{figure}
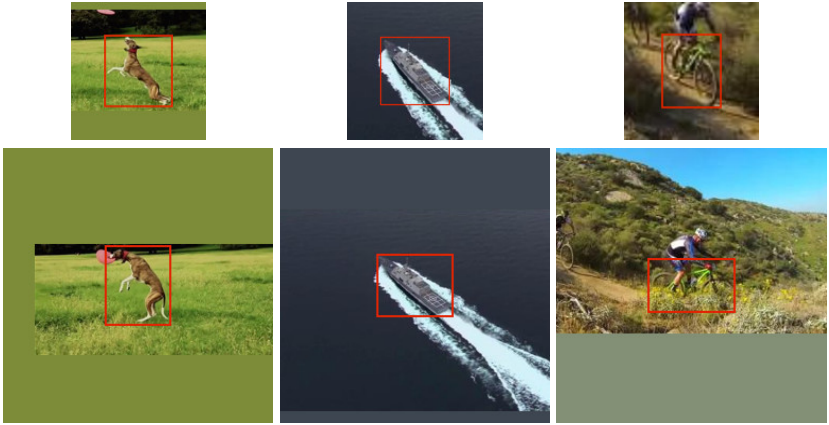

Pairs are obtained from a dataset of annotated videos by extracting exemplar and search images that are centred on the target, as shown in Figure~\ref{fig:pos-pairs}.
The images are extracted from two frames of a video that both contain the object and are at most $T$ frames apart.
The class of the object is ignored during training.
The scale of the object within each image is normalized without corrupting the aspect ratio of the image.
The elements of the score map are considered to belong to a positive example if they are within radius $R$ of the centre (accounting for the stride~$k$ of the network)
\begin{equation}
y[u] = \begin{cases} +1 & \text{if } k \|u - c\| \le R \\ -1 & \text{otherwise} \enspace . \end{cases}
\end{equation}
The losses of the positive and negative examples in the score map are weighted to eliminate class imbalance.

Since our network is fully-convolutional, there is no risk that it learns a bias for the sub-window at the centre.
We believe that it is effective to consider search images centred on the target because it is likely that the most difficult sub-windows, and those which have the most influence on the performance of the tracker, are those adjacent to the target.


Note that since the network is symmetric $f(z, x) = f(x, z)$, it is in fact also fully-convolutional in the exemplar.
While this allows us to use different size exemplar images for different objects in theory, we assume uniform sizes because it simplifies the mini-batch implementation.
However, this assumption could be relaxed in the future.

\subsection{ImageNet Video for tracking}

The 2015 edition of ImageNet Large Scale Visual Recognition Challenge~\cite{ILSVRC15} (ILSVRC) introduced the ImageNet Video dataset as part of the new \emph{object detection from video} challenge.
Participants are required to classify and locate objects from 30 different classes of animals and vehicles.
Training and validation sets together contain almost 4500 videos, with a total of more than one million annotated frames.
This number is particularly impressive if compared to the number of labelled sequences in VOT~\cite{kristan2015visual}, ALOV~\cite{smeulders2014visual} and OTB~\cite{WuLimYang13}, which together total less than 500 videos.
We believe that this dataset should be of extreme interest to the tracking community not only for its vast size, but also because it depicts scenes and objects different to those found in the canonical tracking benchmarks.
For this reason, it can safely be used to train a deep model for tracking without over-fitting to the domain of videos used in these benchmarks.


\subsection{Practical considerations}
\label{sec:considerations}


\subsubsection{Dataset curation}
During training, we adopt exemplar images that are $127 \times 127$ and search images that are $255 \times 255$ pixels.
Images are scaled such that the bounding box, plus an added margin for context, has a fixed area.
More precisely, if the tight bounding box has size $(w, h)$ and the context margin is $p$, then the scale factor~$s$ is chosen such that the area of the scaled rectangle is equal to a constant
\begin{equation}
s (w + 2 p) \times s (h + 2 p) = A \enspace .
\end{equation}
We use the area of the exemplar images $A = 127^2$ and set the amount of context to be half of the mean dimension $p = (w+h) / 4$.
Exemplar and search images for every frame are extracted offline to avoid image resizing during training.
In a preliminary version of this work, we adopted a few heuristics to limit the number of frames from which to extract the training data.
For the experiments of this paper, instead, we have used \emph{all} 4417 videos of ImageNet Video, which account for more than 2 million labelled bounding boxes.

\subsubsection{Network architecture}
The architecture that we adopt for the embedding function~$\varphi$ resembles the convolutional stage of the network of Krizhevsky et al.~\cite{krizhevsky2012imagenet}.
The dimensions of the parameters and activations are given in Table~\ref{tab:architecture}.
Max-pooling is employed after the first two convolutional layers.
ReLU non-linearities follow every convolutional layer except for conv5, the final layer.
During training, batch normalization~\cite{ioffe2015batch} is inserted immediately after every linear layer.
The stride of the final representation is eight.
An important aspect of the design is that no padding is introduced within the network.
Although this is common practice in image classification, it violates the fully-convolutional property of eq.~\ref{eq:commutes-translation}.

\begin{table}[t]
\centering
\caption{Architecture of convolutional embedding function, which is similar to the convolutional stage of the network of Krizhevsky et al.~\cite{krizhevsky2012imagenet}.
The channel map property describes the number of output and input channels of each convolutional layer.
}
\label{tab:architecture}
{
\begin{tabular}{L{8ex} C{10ex} C{14ex} C{8ex} C{16ex} C{16ex} L{8ex}} \hline
& & & & \multicolumn{3}{c}{Activation size} \\
Layer & Support & Chan.\ map & Stride & for exemplar & for search & \multicolumn{1}{c}{chans.} \\ \hline
& & & & $127 \times 127$ & $255 \times 255$ & $\times 3$ \\
conv1 & $11 \times 11$ & $96 \times 3$    & 2 & $59 \times 59$ & $123 \times 123$ & $\times 96$ \\
pool1 & $3 \times 3$   &                  & 2 & $29 \times 29$ & $61 \times 61$   & $\times 96$ \\
conv2 & $5 \times 5$   & $256 \times 48$  & 1 & $25 \times 25$ & $57 \times 57$   & $\times 256$ \\
pool2 & $3 \times 3$   &                  & 2 & $12 \times 12$ & $28 \times 28$   & $\times 256$ \\
conv3 & $3 \times 3$   & $384 \times 256$ & 1 & $10 \times 10$ & $26 \times 26$   & $\times 192$ \\
conv4 & $3 \times 3$   & $384 \times 192$ & 1 & $8 \times 8$   & $24 \times 24$   & $\times 192$ \\
conv5 & $3 \times 3$   & $256 \times 192$ & 1 & $6 \times 6$   & $22 \times 22$   & $\times 128$ \\
\hline
\end{tabular}
}
\end{table}

\subsubsection{Tracking algorithm}
Since our purpose is to prove the efficacy of our fully-convolutional Siamese network and its generalization capability when trained on ImageNet Video, we use an extremely simplistic algorithm to perform tracking.
Unlike more sophisticated trackers, we do not update a model or maintain a memory of past appearances, we do not incorporate additional cues such as optical flow or colour histograms, and we do not refine our prediction with bounding box regression.
Yet, despite its simplicity, the tracking algorithm achieves surprisingly good results when equipped with our offline-learnt similarity metric.
Online, we do incorporate some elementary temporal constraints: we only search for the object within a region of approximately four times its previous size, and a cosine window is added to the score map to penalize large displacements.
Tracking through scale space is achieved by processing several scaled versions of the search image.
Any change in scale is penalized and updates of the current scale are damped.

%% file: art/pairs/figure.tex
\begin{tabular}{ccc} 

\includegraphics[width=18mm]{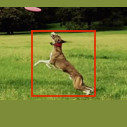}
& \includegraphics[width=18mm]{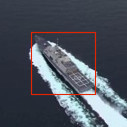}
& \includegraphics[width=18mm]{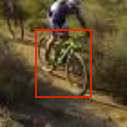}
\\
\includegraphics[width=36mm]{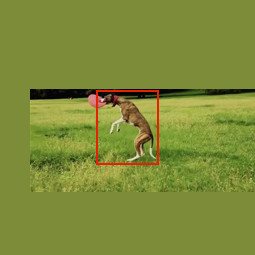}
& \includegraphics[width=36mm]{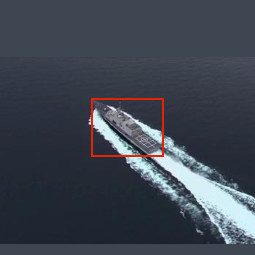}
& \includegraphics[width=36mm]{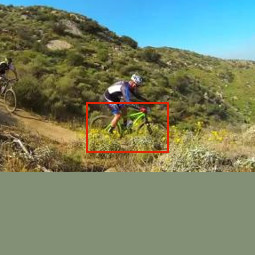}

\end{tabular}

%% file: related.tex
Several recent works have sought to train Recurrent Neural Networks (RNNs) for the problem of object tracking.
Gan et al.~\cite{gan2015first} train an RNN to predict the absolute position of the target in each frame and Kahou et al.~\cite{kahou2015ratm} similarly train an RNN for tracking using a differentiable attention mechanism.
These methods have not yet demonstrated competitive results on modern benchmarks, however it is certainly a promising avenue for future research.
We remark that an interesting parallel can be drawn between this approach and ours, by interpreting a Siamese network as an unrolled RNN that is trained and evaluated on sequences of length two.
Siamese networks could therefore serve as strong initialization for a recurrent model.

Denil et al.~\cite{denil2012learning} track objects with a particle filter that uses a learnt distance metric to compare the current appearance to that of the first frame.
However, their distance metric is vastly different to ours.
Instead of comparing images of the entire object, they compute distances between fixations (foveated glimpses of small regions within the object's bounding box).
To learn a distance metric, they train a Restricted Boltzmann Machine (RBM) and then use the Euclidean distance between hidden activations for two fixations.
Although RBMs are unsupervised, they suggest training the RBM on random fixations within centred images of the object to detect.
This must either be performed online or in an offline phase with knowledge of the object to track.
While tracking an object, they learn a stochastic policy for choosing fixations which is specific to that object, using uncertainty as a reward signal.
Besides synthetic sequences of MNIST digits, this method has only been demonstrated qualitatively on problems of face and person tracking.

While it is infeasible to train a deep conv-net from scratch for each new video, several works have investigated the feasibility of fine-tuning from pre-trained parameters at test time.
SO-DLT~\cite{wang2015transferring} and MDNet~\cite{nam2015learning} both train a convolutional network for a similar detection task in an offline phase, then at test-time use SGD to learn a detector with examples extracted from the video itself as in the conventional tracking-as-detector-learning paradigm.
These methods cannot operate at frame-rate due to the computational burden of evaluating forward and backward passes on many examples.
An alternative way to leverage conv-nets for tracking is to apply traditional shallow methods using the internal representation of a pre-trained convolutional network as features.
While trackers in this style such as DeepSRDCF~\cite{danelljan2015convolutional}, Ma et al.~\cite{ma2015hierarchical} and FCNT~\cite{wang2015visual} have achieved strong results, they have been unable to achieve frame-rate operation due to the relatively high dimension of the conv-net representation.

Concurrently with our own work, some other authors have also proposed using conv-nets for object tracking by learning a function of pairs of images.
Held et al.~\cite{held2016learning} introduce GOTURN, in which a conv-net is trained to regress directly from two images to the location in the second image of the object shown in the first image.
Predicting a rectangle instead of a position has the advantage that changes in scale and aspect ratio can be handled without resorting to exhaustive evaluation.
However, a disadvantage of their approach is that it does not possess intrinsic invariance to translation of the second image.
This means that the network must be shown examples in all positions, which is achieved through considerable dataset augmentation.
Chen et al.~\cite{chen2016once} train a network that maps an exemplar and a larger search region to a response map.
However, their method also lacks invariance to translation of the second image since the final layers are fully-connected.
Similarly to Held et al., this is inefficient because the training set must represent all translations of all objects.
Their method is named YCNN for the Y shape of the network.
Unlike our approach, they cannot adjust the size of the search region dynamically after training.
Tao et al.~\cite{tao2016siamese} propose to train a Siamese network to identify candidate image locations that match the initial object appearance, dubbing their method SINT (Siamese INstance search Tracker).
In contrast to our approach, they do not adopt an architecture which is fully-convolutional with respect to the search image.
Instead, at test time, they sample bounding boxes uniformly on circles of varying radius as in Struck~\cite{hare2011struck}.
Moreover, they incorporate optical flow and bounding box regression to improve the results.
In order to improve the computational speed of their system, they employ Region of Interest (RoI) pooling to efficiently examine many overlapping sub-windows.
Despite this optimization, at 2 frames per second, the overall system is still far from being real-time.

All of the competitive methods above that train on video sequences (MDNet~\cite{nam2015learning}, SINT~\cite{tao2016siamese}, GOTURN~\cite{held2016learning}), use training data belonging to the same ALOV/OTB/VOT domain used by the benchmarks.
This practice has been forbidden in the VOT challenge due to concerns about over-fitting to the scenes and objects in the benchmark.
Thus an important contribution of our work is to demonstrate that a conv-net can be trained for effective object tracking without using videos from the same distribution as the testing set.

%% file: experiments.tex
\subsection{Implementation details}

\subsubsection{Training.}
The parameters of the embedding function are found by minimizing eq.~\ref{eq:training} with straightforward SGD using MatConvNet~\cite{vedaldi15matconvnet}.
The initial values of the parameters follow a Gaussian distribution, scaled according to the improved Xavier method~ \cite{he2015delving}.
Training is performed over 50 epochs, each consisting of 50,000 sampled pairs (according to sec.~\ref{sec:training}).
The gradients for each iteration are estimated using mini-batches of size 8, and the learning rate is annealed geometrically at each epoch from $10^{-2}$ to $10^{-5}$.

\subsubsection{Tracking.}
As mentioned earlier, the online phase is deliberately minimalistic. The embedding~$\varphi(z)$ of the initial object appearance is computed once, and is compared convolutionally to sub-windows of the subsequent frames.
We found that updating (the feature representation of) the exemplar online through simple strategies, such as linear interpolation, does not gain much performance and thus we keep it fixed.
We found that upsampling the score map using bicubic interpolation, from $17\times17$ to $272\times272$, results in more accurate localization since the original map is relatively coarse.
To handle scale variations, we also search for the object over five scales $1.025^{\{-2,-1,0,1,2\}}$, and update the scale by linear interpolation with a factor of 0.35 to provide damping.

In order to make our experimental results reproducible, we share training and tracking code, together with the scripts to generate the curated dataset at \url{http://www.robots.ox.ac.uk/~luca/siamese-fc.html}.
On a machine equipped with a single NVIDIA GeForce GTX Titan X and an Intel Core i7-4790K at 4.0GHz, our full online tracking pipeline operates at 86 and 58 frames-per-second, when searching respectively over 3 and 5 scales.

\subsection{Evaluation}
We evaluate two variants of our simplistic tracker: SiamFC (Siamese Fully-Convolutional) and SiamFC-3s, which searches over 3 scales instead of 5.


\subsection{The OTB-13 benchmark}
\begin{figure}[t!]
    \centering
    \begin{subfigure}[t]{0.32\textwidth}
        \centering
        \includegraphics[height=.7\linewidth]{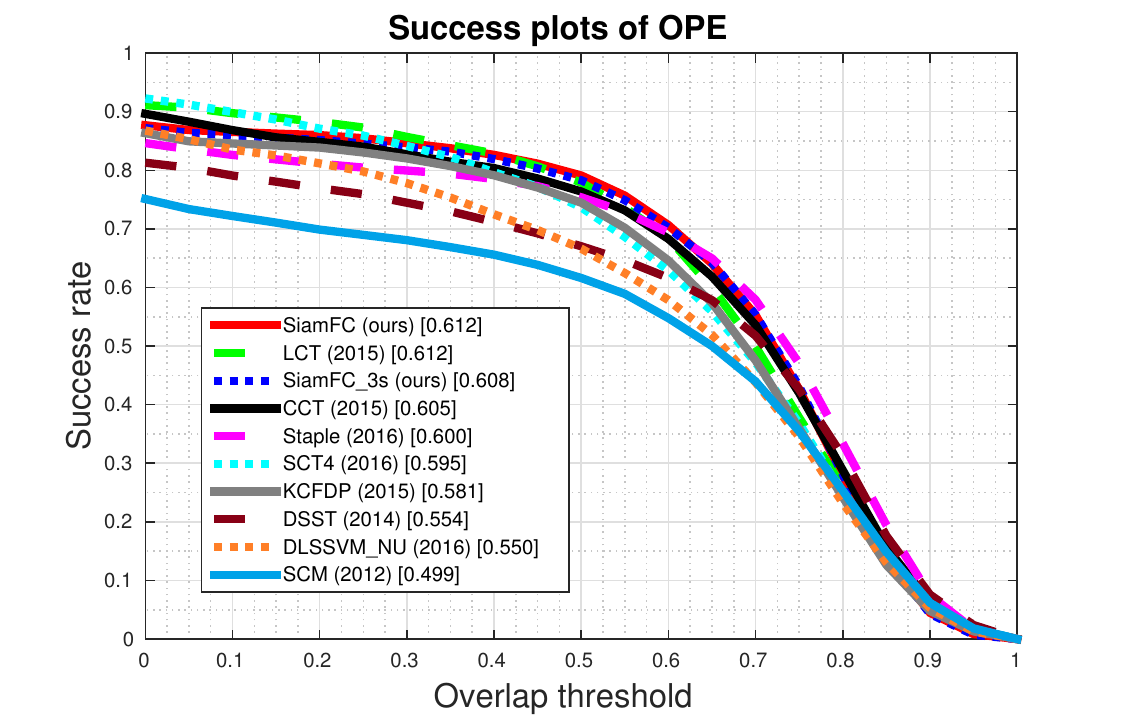}
    \end{subfigure}%
    ~
    \begin{subfigure}[t]{0.32\textwidth}
        \centering
        \includegraphics[height=.7\linewidth]{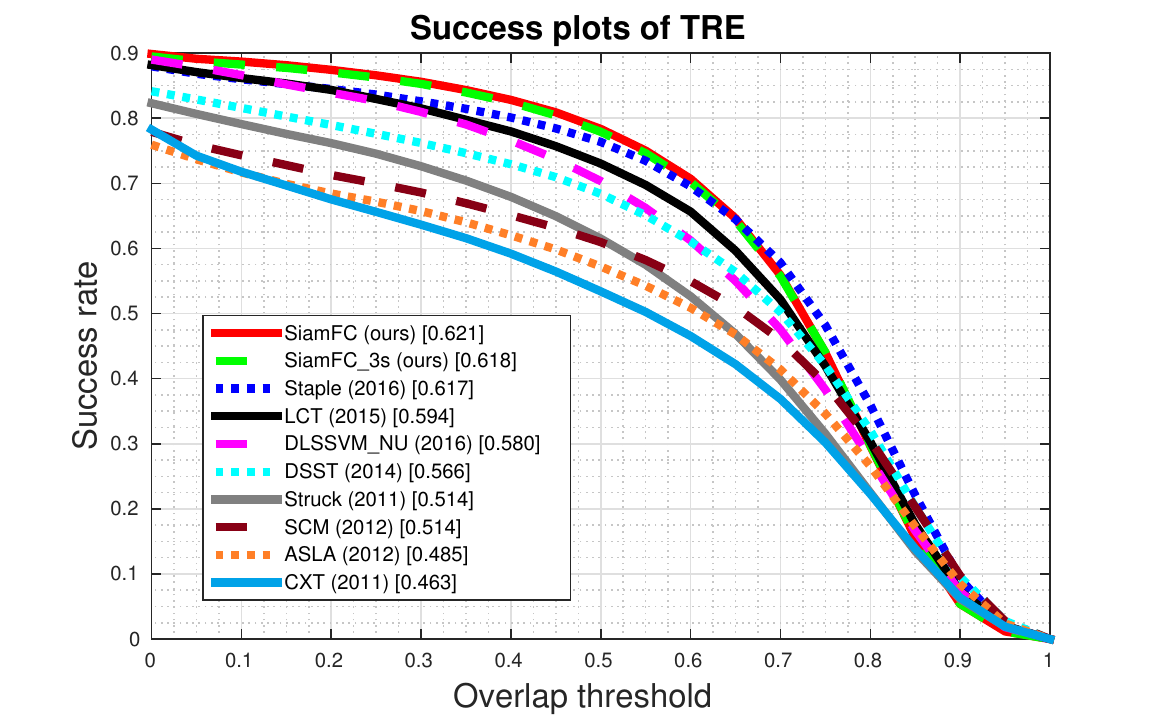}
    \end{subfigure}
    ~
    \begin{subfigure}[t]{0.32\textwidth}
        \centering
        \includegraphics[height=.7\linewidth]{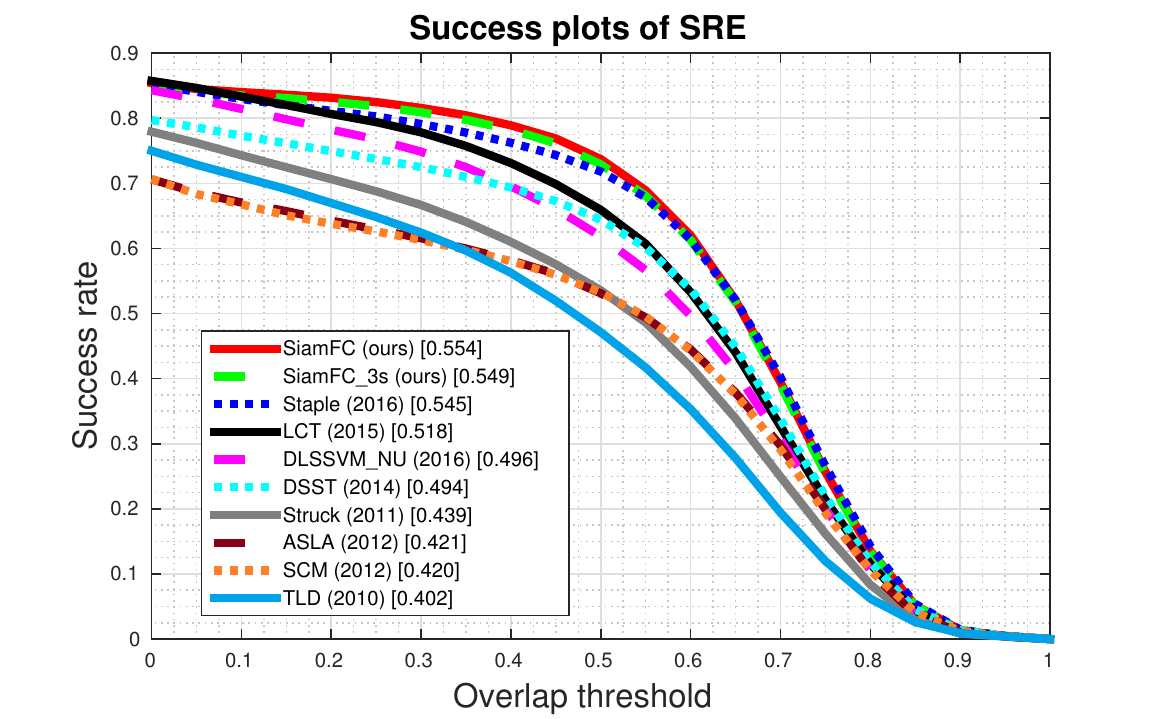}
    \end{subfigure}
    \caption{Success plots for OPE (one pass evaluation), TRE (temporal robustness evaluation) and SRE (spatial robustness evaluation) of the OTB-13~\cite{WuLimYang13} benchmark. The results of CCT, SCT4 and KCFDP were only available for OPE at the time of writing.}
    \label{fig:OTB}
\end{figure}

The OTB-13~\cite{WuLimYang13} benchmark considers the average per-frame \textit{success rate} at different thresholds: a tracker is successful in a given frame if the intersection-over-union (IoU) between its estimate and the ground-truth is above a certain threshold.
Trackers are then compared in terms of area under the curve of success rates for different values of this threshold.
In addition to the trackers reported by~\cite{WuLimYang13}, in Figure~\ref{fig:OTB} we also compare against seven more recent state-of-the-art trackers presented in the major computer vision conferences and that can run at frame-rate speed: Staple~\cite{bertinetto2015staple}, LCT~\cite{ma2015long}, CCT~\cite{zhu2015collaborative}, SCT4~\cite{choi2016visual}, DLSSVM\_NU~\cite{Ning_2016_CVPR}, DSST~\cite{danelljan2014accurate} and KCFDP~\cite{huang12enable}.
Given the nature of the sequences, for this benchmark only we convert 25\% of the pairs to grayscale during training.
All the other hyper-parameters (for training and tracking) are fixed.


\subsection{The VOT benchmarks}
For our experiments, we use the latest stable version of the Visual Object Tracking (VOT) toolkit (tag \texttt{vot2015-final}), which evaluates trackers on sequences chosen from a pool of 356, selected so that seven different challenging situations are well represented.
Many of the sequences were originally presented in other datasets (\eg ALOV~\cite{smeulders2014visual} and OTB~\cite{WuLimYang13}).
Within the benchmark, trackers are automatically re-initialized five frames after failure, which is deemed to have occurred when the IoU between the estimated bounding box and the ground truth becomes zero.

\subsubsection{VOT-14 results.}
\begin{figure}[t]
\centering
\includegraphics[width=.98\textwidth]{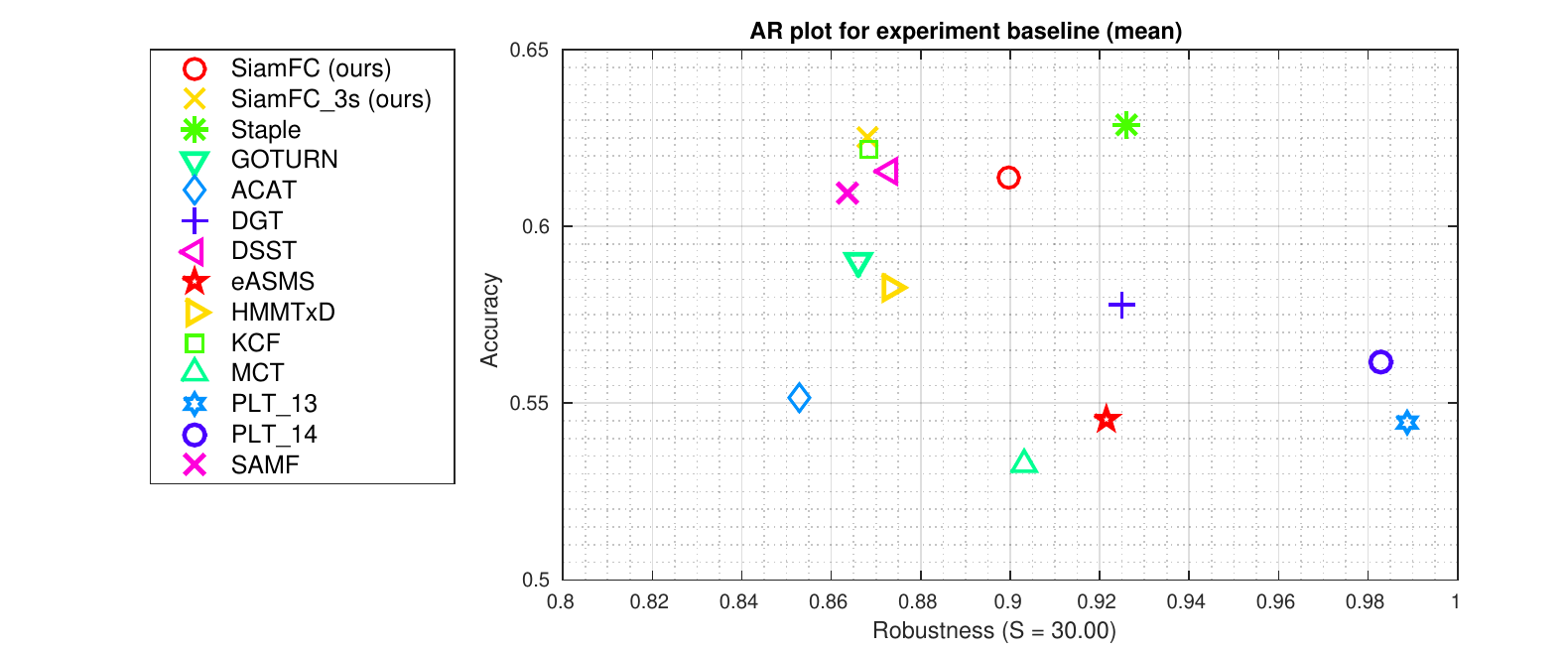}
\caption{VOT-14 Accuracy-robustness plot. Best trackers are closer to the top-right corner.}
\label{fig:vot14-AR}
\end{figure}

We compare our method SiamFC (and the variant SiamFC-3s) against the best 10 trackers that participated in the 2014 edition of the VOT challenge~\cite{lirisvisual}. We  also include Staple~\cite{bertinetto2015staple} and GOTURN~\cite{held2016learning}, two recent real-time trackers presented respectively at CVPR 2016 and ECCV 2016.
Trackers are evaluated according to two measures of performance: \emph{accuracy} and \emph{robustness}.
The former is calculated as the average IoU, while the latter is expressed in terms of the total number of failures.
These give insight into the behaviour of a tracker.
Figure~\ref{fig:vot14-AR} shows the Accuracy-Robustness plot, where the best trackers are closer to the top-right corner.

\subsubsection{VOT-15 results.}
\begin{figure}[t]
\centering
\includegraphics[width=0.98\textwidth]{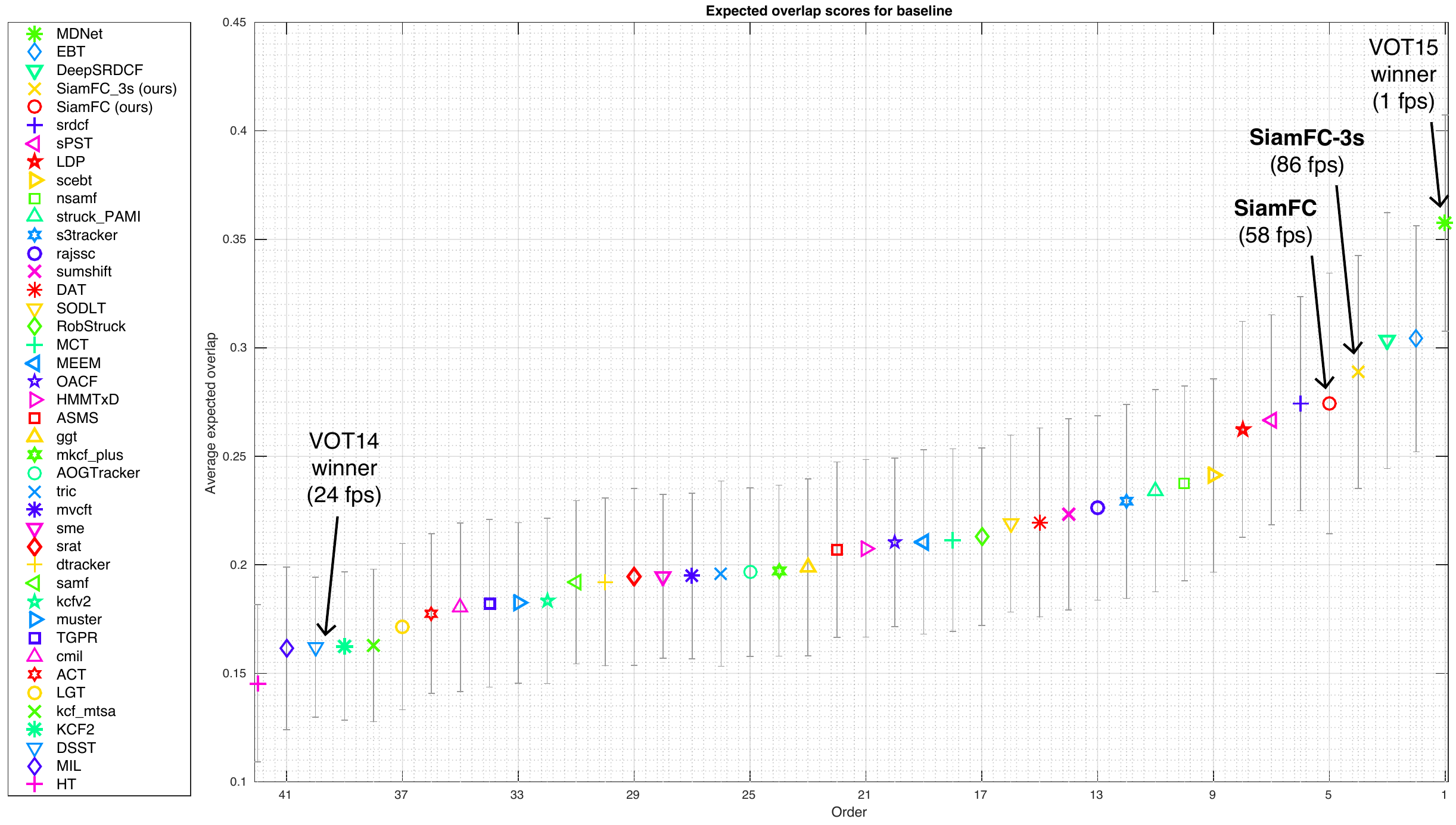}
\caption{VOT-15 ranking in terms of expected average overlap. Only the best 40 results have been reported.}
\label{fig:vot15-overlap}
\end{figure}

We also compare our method against the 40 best participants in the 2015 edition~\cite{kristan2015visual}.
In this case, the raw scores of accuracy and number of failures are used to compute the \emph{expected average overlap measure}, which represents the average IoU with no re-initialization following a failure.
Figure~\ref{fig:vot15-overlap} illustrates the final ranking in terms of expected average overlap, while Table~\ref{tab:vot2015-raw} reports scores and speed of the 15 highest ranked trackers of the challenge.

\begin{table}[t]
\centering
\caption{Raw scores, overlap and reported speed for our proposed method and the best 15 performing trackers of the VOT-15 challenge. Where available, we compare with the speed reported by the authors, otherwise (*) we report the values from the VOT-15 results~\cite{kristan2015visual} in EFO units, which roughly correspond to fps (\eg the speed of the NCC tracker is 140 fps and 160 EFO).
}
\label{tab:vot2015-raw}
\begin{tabular}{L{3.5cm} L{2.1cm} L{2.1cm} L{2.1cm} R{0.7cm} L{0.7cm}}\hline
Tracker & accuracy & \# failures & overlap & \multicolumn{2}{l}{speed (fps)} \\ \hline
MDNet~\cite{nam2015learning} & 0.5620 & 46 & 0.3575 & 1 \\
EBT~\cite{zhu2015tracking} & 0.4481 & 49 & 0.3042 & 5 \\
DeepSRDCF~\cite{danelljan2015convolutional} & 0.5350 & 60 & 0.3033 & $< 1$ & *  \\
\textbf{SiamFC-3s} (ours) & 0.5335 & 84 & 0.2889 & \textbf{86}\\
\textbf{SiamFC} (ours) & 0.5240 & 87 & 0.2743 & 58\\
SRDCF~\cite{danelljan2015learning} & 0.5260 & 71 & 0.2743 & 5 \\
sPST~\cite{hua2015online} & 0.5230 & 85 & 0.2668 & 2 \\
LDP~\cite{kristan2015visual} & 0.4688 & 78 & 0.2625 & 4 & * \\
SC-EBT~\cite{wang2014ensemble} & 0.5171 & 103 & 0.2412 & -- \\
NSAMF~\cite{li2014scale} & 0.5027 & 87 & 0.2376 & 5 & * \\
StruckMK~\cite{hare2011struck} & 0.4442 & 90 & 0.2341 & 2  \\
S3Tracker~\cite{li2015nus} & 0.5031 & 100 & 0.2292 & 14 & *  \\
RAJSSC~\cite{kristan2015visual} & 0.5301 & 105 & 0.2262 & 2 & *  \\
SumShift~\cite{li2015nus} & 0.4888 & 97 & 0.2233 & 17 & * \\
DAT~\cite{possegger2015defense} & 0.4705 & 113 & 0.2195 & 15 \\
SO-DLT~\cite{wang2015transferring} & 0.5233 & 108 & 0.2190 & 5 \\
\hline
\end{tabular}
\end{table}

\subsubsection{VOT-16 results.}
At the time of writing, the results of the 2016 edition were not available.
However, to facilitate an early comparison with our method, we report our scores.
For SiamFC and SiamFC-3s we obtain, respectively, an overall expected overlap (average between the \textit{baseline} and \textit{unsupervised} experiments) of 0.3876 and 0.4051.
Please note that these results are different from the VOT-16 report, as our entry in the challenge was a preliminary version of this work.

\subsubsection*{}Despite its simplicity, our method improves over recent state-of-the-art real-time trackers (Figures~\ref{fig:OTB} and~\ref{fig:vot14-AR}).
Moreover, it outperforms most of the best methods in the challenging VOT-15 benchmark, while being the only one that achieves frame-rate speed (Figure~\ref{fig:vot15-overlap} and Table~\ref{tab:vot2015-raw}).
These results demonstrate that the expressiveness of the similarity metric learnt by our fully-convolutional Siamese network on ImageNet Video \emph{alone} is enough to achieve very strong results, comparable or superior to recent state-of-the-art methods, which often are several orders of magnitude slower.
We believe that considerably higher performance could be obtained by augmenting the minimalist online tracking pipeline with the methods often adopted by the tracking community (\eg model update, bounding-box regression, fine-tuning, memory).

\begin{figure}[h]
\centering
\includegraphics[width=0.99\textwidth]{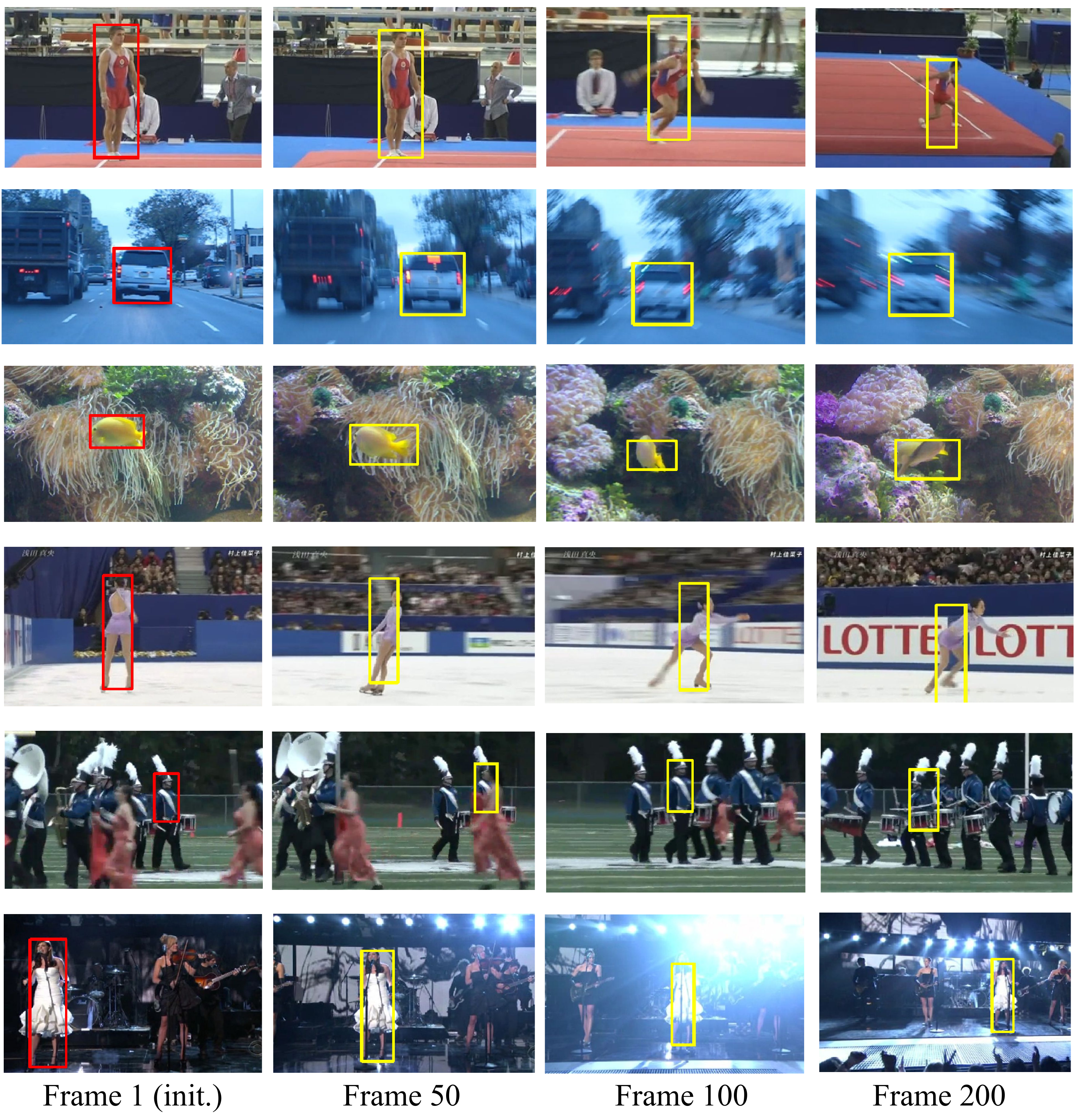}
\caption{Snapshots of the simple tracker described in Section~\ref{sec:considerations} equipped with our proposed fully-convolutional Siamese network trained from scratch on ImageNet Video.
Our method does not perform any model update, so it uses only the first frame to compute $\varphi(z)$.
Nonetheless, it is surprisingly robust to a number of challenging situations like motion blur (row 2), drastic change of appearance (rows 1, 3 and 4), poor illumination (row 6) and scale change (row 6).
On the other hand, our method is sensitive to scenes with confusion (row 5), arguably because the model is never updated and thus the cross-correlation gives a high scores for all the windows that are similar to the first appearance of the target.
All sequences come from the VOT-15 benchmark: \emph{gymnastics1}, \emph{car1}, \emph{fish3}, \emph{iceskater1}, \emph{marching}, \emph{singer1}.
The snapshots have been taken at fixed frames (1, 50, 100 and 200) and the tracker is never re-initialized.}
\label{fig:tracking_snapshots}
\end{figure}

\subsection{Dataset size}
Table~\ref{tab:dataset-size} illustrates how the size of the dataset used to train the Siamese network greatly influences the performance.
The expected average overlap (measured on VOT-15) steadily improves from 0.168 to 0.274 when increasing the size of the dataset from 5\% to 100\%.
This finding suggests that using a larger video dataset could increase the performance even further.
In fact, even if 2 million supervised bounding boxes might seem a huge number, it should not be forgotten that they still belong to a relatively moderate number of videos, at least compared to the amount of data normally used to train conv-nets.

\begin{table}[t]
\centering
\caption{Effects of using increasing portions of the ImageNet Video dataset on tracker's performance.
}
\label{tab:dataset-size}
\begin{tabular}{C{1.9cm} C{1.9cm} C{1.9cm} C{1.9cm} C{1.9cm} C{1.9cm}} \hline
Dataset (\%) & \# videos & \# objects & accuracy & \# failures & expected avg. overlap\\ \hline
2 & 88 & 60k & 0.484 & 183 & 0.168 \\
4 & 177 & 110k & 0.501 & 160 & 0.192 \\
8 & 353 & 190k & 0.484 & 142 & 0.193 \\
16 & 707 & 330k & 0.522 & 132 & 0.219 \\
32 & 1413 & 650k & 0.521 & 117 & 0.234 \\
100 & 4417 & 2m & \textbf{0.524} & \textbf{87} & \textbf{0.274} \\
\hline
\end{tabular}
\end{table}